\documentclass{article}
\usepackage[nonanonymous]{neurips_2026}
\usepackage[utf8]{inputenc}
\usepackage[T1]{fontenc}
\usepackage{hyperref}
\usepackage{url}
\usepackage{booktabs}
\usepackage{amsfonts}
\usepackage{nicefrac}
\usepackage{microtype}
\usepackage{xcolor}
\usepackage{amsmath,amssymb}
\usepackage{array}
\usepackage{makecell}
\usepackage{multirow}
\usepackage{colortbl}
\usepackage{graphicx}
\definecolor{headerbg}{gray}{0.93}
\usepackage{algorithm}
\usepackage{algorithmic}
\usepackage{stfloats}
\usepackage{comment}
\usepackage{etoolbox}

\AtBeginEnvironment{equation}{\small}
\AtBeginEnvironment{align}{\small}
\AtBeginEnvironment{gather}{\small}
\AtBeginEnvironment{multline}{\small}

\title{Multi-Rate Mixture of Experts for Accelerating Liquid Neural Network Training}

\author{
Shilong Zong\textsuperscript{1}
\quad
Almuatazbellah Boker\textsuperscript{2}
\quad
Hoda Eldardiry\textsuperscript{1}
\\[0.5em]
\textsuperscript{1}Department of Computer Science, Virginia Tech
\\
\textsuperscript{2}Department of Electrical and Computer Engineering, Virginia Tech
\\
Blacksburg, VA 24061 USA
\\[0.5em]
\texttt{shilongz@vt.edu}
\quad
\texttt{boker@vt.edu}
\quad
\texttt{hdardiry@vt.edu}
}

\date{April 2026}

\begin{document}

\nolinenumbers
\maketitle

\maketitle
\begin{abstract}
Multivariate time-series data often exhibit complex temporal dependencies, irregular sampling, and heterogeneous dynamics across multiple time scales, making accurate sequence modeling particularly challenging. Traditional recurrent neural networks (RNNs), such as Long Short-Term Memory (LSTM) networks, operate in discrete time and may struggle to effectively capture continuous and irregular temporal behaviors. Liquid Neural Networks (LNNs) address some of these limitations through continuous-time dynamics, but standard LNN architectures typically rely on a single dynamical system, limiting their ability to model heterogeneous temporal patterns.
To address these challenges, we propose a Multi-Rate Mixture-of-Experts (MR-MoE) framework built on top of Liquid Neural Networks. In the proposed architecture, multiple LNN-based experts operate at distinct time scales, enabling the model to explicitly separate fast-changing dynamics from slow-evolving temporal trends. A gating network further enables adaptive expert specialization based on input conditions. In addition, we incorporate both feature-level and temporal attention mechanisms to improve robustness, interpretability, and long-range
dependency modeling. Feature-level attention suppresses noisy or irrelevant variables, while temporal attention selectively focuses on informative historical states.
We evaluate the proposed framework on a complex multivariate time-series prediction task and compare it against strong baselines, including LSTM, monolithic LNN, and standard MoE models. Experimental results demonstrate that the proposed MR-MoE framework consistently achieves improved AUROC and AUPRC performance while maintaining favorable computational efficiency. These results highlight the effectiveness of combining continuous-time dynamics, multi-scale expert decomposition, and adaptive attention mechanisms for time-series modeling.
\end{abstract}

\section{Introduction}

Multivariate time-series data are widely encountered in many real-world applications and are often characterized by complex temporal dependencies, irregular sampling, and significant noise. Developing models that can effectively capture these characteristics remains a fundamental challenge in time-series prediction~\cite{che2018recurrent,moor2023predicting}.

Recurrent neural networks (RNNs), such as Long Short-Term Memory (LSTM), have been widely applied to sequential and time-series modeling tasks. However, these models operate in discrete time and often struggle to capture long-range dependencies and irregular temporal dynamics. Recent studies comparing LNNs and RNNs have shown that continuous-time dynamics can provide advantages in temporal modeling, memory efficiency, and generalization ability~\cite{zong2026lnnreview}. Liquid Neural Networks (LNNs) address some of these limitations by modeling continuous-time dynamics through differential equations~\cite{chen2018neural,hasani2021liquid}. Each differential equation is characterized by a time constant, which controls the rate of state evolution. The notion of “liquid” refers to the adaptability of these time constants, allowing the model to better capture underlying temporal dynamics~\cite{rubanova2019latent,kidger2020neural}.

Nevertheless, standard LNNs typically rely on a single dynamical system, meaning that a single set of differential equations governs the evolution of the hidden state across all inputs. While this provides a compact representation, it limits the model's ability to capture heterogeneous patterns and multi-scale temporal structures, where different underlying processes may evolve at distinct time scales. This limitation motivates the need for structured models that can explicitly separate and efficiently model different temporal scales.

In addition, LNNs often incur higher computational costs due to the need for numerical integration of differential equations, which can become a bottleneck when scaling to large datasets or long time horizons. This makes efficient modeling of large-scale time-series particularly challenging.

To address these challenges, we propose a structured framework that combines multi-scale modeling, expert specialization, and attention mechanisms. Specifically, we first introduce a Mixture-of-Experts (MoE) framework based on Liquid Neural Networks, where multiple experts improve representation capacity and capture diverse temporal patterns through expert specialization~\cite{shazeer2017outrageously}. Building upon this idea, we further propose a Multi-Rate Mixture-of-Experts (MR-MoE) architecture, where experts operate at distinct time scales to explicitly separate fast-changing dynamics from slow-evolving temporal trends.

Furthermore, we incorporate both feature-level and temporal attention mechanisms into the proposed framework~\cite{bahdanau2015neural,choi2016retain}. Feature-level attention identifies the most relevant input variables and suppresses noisy or irrelevant features, while temporal attention selectively focuses on informative historical states to improve long-range dependency modeling. Existing LNN-based models typically focus on either attention mechanisms or expert specialization independently~\cite{hasani2021liquid,shazeer2017outrageously}. In contrast, the proposed framework integrates continuous-time dynamics, expert specialization, multi-scale modeling, and adaptive attention within a unified architecture. The proposed Multi-Rate-MoE framework is the only architecture that combines both attention mechanisms and MoE-based expert decomposition simultaneously.

To validate the effectiveness of the proposed framework, we evaluate the model on a complex multivariate time-series prediction task. Experimental results demonstrate that the proposed method consistently outperforms strong baselines, including LSTM, monolithic LNN, and standard MoE models.

    \paragraph{Contributions.}
The main contributions of this work are summarized as follows:

\begin{itemize}

\item We introduce a Mixture-of-Experts (MoE) framework based on Liquid Neural Networks (LNNs) for time-series modeling. By decomposing the model into multiple experts, the framework improves representation capacity and enables different experts to capture diverse temporal patterns. Compared to a single monolithic model, the MoE structure provides better flexibility, specialization, and robustness.

\item We further propose a Multi-Rate Mixture-of-Experts (MR-MoE) architecture, where different experts operate at distinct time scales. This design enables the model to explicitly separate fast-changing dynamics from slow-evolving trends, improving the modeling of heterogeneous temporal behaviors while reducing interference between different temporal processes.

\item We incorporate both feature-level and temporal attention mechanisms into the MR-MoE framework. Feature-level attention improves robustness by identifying important input variables and suppressing noisy or irrelevant features, while temporal attention enhances long-range dependency modeling by selectively focusing on informative historical states. The combination of MR-MoE and attention mechanisms further improves both prediction performance and model interpretability.

\item We demonstrate the effectiveness of the proposed approach on sepsis prediction. Experimental results show consistent improvements over strong baselines, including LSTM, monolithic LNN, and MoE models.

\end{itemize}

\section{Method}

    In this work, we progressively develop our model from standard sequence modeling to a structured multi-scale framework with attention mechanisms. Specifically, we begin with an LSTM baseline, extend it to continuous-time modeling via Liquid Neural Networks (LNNs), introduce a Mixture-of-Experts (MoE) architecture for specialization, incorporate the singular perturbation method to model multi-time-scale dynamics, and enhance the model with both temporal and feature-level attention mechanisms.

    \subsection{LSTM Baseline}
As a baseline, we consider Long Short-Term Memory (LSTM) networks~\cite{hochreite1997long} for sequence modeling. Given an input sequence $u(t)$, the hidden state is updated as Eq.~\ref{eq1}:

\begin{equation}
\label{eq1}
h_t = \mathrm{LSTM}(h_{t-1}, u_t)
\end{equation}

where $h_t$ denotes the hidden state at time step $t$, and $u_t$ represents the input at time step $t$. The prediction is obtained via Eq.~\ref{eq2}:

\begin{equation}
\label{eq2}
y(t) = W h_t
\end{equation}

where $W$ is a learnable linear projection matrix that maps the hidden state $h_t$ to the output space. While LSTMs are effective for sequential data, they operate in discrete time and may struggle to capture complex temporal dynamics in irregular time-series data~\cite{rubanova2019latent,kidger2020neural}.

\subsection{Liquid Neural Networks}
To better model continuous-time dynamics, we adopt Liquid Neural Networks (LNNs). The hidden state evolves according to Eq.~\ref{eq3}:

\begin{equation}
\label{eq3}
\frac{dx(t)}{dt} = f(x(t), u(t); \theta)
\end{equation}

where $x(t)$ denotes the hidden state, $u(t)$ is the input sequence at time $t$, $f(\cdot)$ represents a nonlinear dynamical function parameterized by $\theta$, and $\theta$ denotes the learnable model parameters. The continuous dynamics are discretized as Eq.~\ref{eq4}:

\begin{equation}
\label{eq4}
x(t+\Delta t) = x(t) + \frac{\Delta t}{\tau} f(x(t), u(t))
\end{equation}

where $\Delta t$ denotes the discretization step size and $\tau$ is a time constant controlling the evolution speed. The output is computed as Eq.~\ref{eq5}:

\begin{equation}
\label{eq5}
y(t) = C x(t)
\end{equation}

where $C$ is a learnable readout matrix that maps the hidden state to the output space. LNNs provide a natural framework for modeling continuous-time processes and irregular sampling patterns~\cite{chen2018neural,hasani2021liquid,rubanova2019latent,kidger2020neural}.

\subsection{Mixture of Experts (MoE)}
To improve model expressiveness, we introduce a Mixture-of-Experts (MoE) architecture~\cite{jacobs1991adaptive,shazeer2017outrageously}. We consider $K$ experts, where each expert is implemented as a Liquid Neural Network (LNN) with its own parameters. Each expert follows the LNN dynamics described above and produces an output $y_k(t)$ based on its hidden state. This allows different experts to capture distinct temporal patterns. A gating network determines the contribution of each expert through Eq.~\ref{eq6}, where $z(t)$ denotes the input representation to the gating network, $g(\cdot)$ is the gating function, and $\phi$ represents the learnable parameters of the gating network.

\begin{equation}
\label{eq6}
\pi(t) = \mathrm{softmax}(g(z(t); \phi))
\end{equation}

The final prediction is computed as a weighted combination of expert outputs in Eq.~\ref{eq7}:

\begin{equation}
\label{eq7}
y(t) = \sum_{k=1}^{K} \pi_k(t) y_k(t)
\end{equation}

where $\pi_k(t)$ denotes the gating weight assigned to expert $k$ at time $t$.

    \subsection{Multi-Rate-Mixture of Experts (MR-MoE)}
Different temporal processes in time-series data often evolve at distinct rates. Some patterns change rapidly over short time intervals, while others evolve gradually over longer horizons. Standard MoE models do not explicitly separate these dynamics, which may limit their ability to efficiently capture heterogeneous temporal behaviors. To address this limitation, we introduce a multi-rate structure where experts operate at different time scales. Each expert is associated with a distinct time constant as Eq.~\ref{eq8}:

\begin{equation}
\label{eq8}
\tau_1 \ll \tau_2 \ll \cdots \ll \tau_K
\end{equation}

where smaller $\tau_k$ corresponds to faster dynamics and larger $\tau_k$ captures slower temporal trends. Following singular perturbation theory~\cite{kokotovic1999singular}, fast experts are approximated using quasi-steady-state mappings as Eq.~\ref{eq9}:

\begin{equation}
\label{eq9}
x_k(t) \approx h_k(x_{\text{slow}}(t), u(t))
\end{equation}

where $x_{\text{slow}}(t)$ denotes the slow-scale hidden states and $h_k(\cdot)$ represents the quasi-steady-state mapping for expert $k$. Slow experts evolve according to continuous dynamics in Eq.~\ref{eq10}:

\begin{equation}
\label{eq10}
\frac{dx_k(t)}{dt} = f_k(x_k(t), u(t))
\end{equation}

where $f_k(\cdot)$ denotes the nonlinear dynamics associated with expert $k$. The final output remains as Eq.~\ref{eq11}:

\begin{equation}
\label{eq11}
y(t) = \sum_{k=1}^{K} \pi_k(t) y_k(t)
\end{equation}

This design enables efficient modeling of both fast-changing and slow-evolving temporal patterns.

\subsection{Multi-Rate-MoE with Attention Mechanisms (MR-MoE-Attention)}
To further improve performance and interpretability, we incorporate both feature-level and temporal attention mechanisms.

\paragraph{Feature-level Attention.}
Given input $u(t) \in \mathbb{R}^d$, we compute feature importance scores using Eq.~\ref{eq12}:

\begin{equation}
\label{eq12}
e(t) = f_{\text{att}}(u(t))
\end{equation}

where $f_{\text{att}}(\cdot)$ denotes the feature-attention network. The attention weights are normalized using softmax in Eq.~\ref{eq13}:

\begin{equation}
\label{eq13}
\beta_j(t) = \frac{\exp(e_j(t))}{\sum_{m=1}^{d} \exp(e_m(t))}
\end{equation}

where $\beta_j(t)$ denotes the attention weight assigned to feature $j$ at time $t$. The input is then reweighted as Eq.~\ref{eq14}:

\begin{equation}
\label{eq14}
\tilde{u}(t) = \beta(t) \odot u(t)
\end{equation}

where $\beta(t) \in \mathbb{R}^d$ denotes the feature attention vector and $\odot$ represents element-wise multiplication. This mechanism acts as a soft feature selection layer, suppressing irrelevant variables and improving the signal-to-noise ratio.

\paragraph{Temporal Attention.}
For each expert $k$, we compute attention weights over historical hidden states using Eq.~\ref{eq15}:

\begin{equation}
\label{eq15}
\alpha_k(t,i) = \frac{\exp(q_k(t)^\top x_k(i))}{\sum_j \exp(q_k(t)^\top x_k(j))}
\end{equation}

where $\alpha_k(t,i)$ denotes the attention weight assigned to the historical hidden state $x_k(i)$ for expert $k$, and $q_k(t)$ represents the query vector at time $t$ used to measure similarity between current and historical states. The context vector is computed as Eq.~\ref{eq16}:

\begin{equation}
\label{eq16}
h_k(t) = \sum_i \alpha_k(t,i) x_k(i)
\end{equation}

where $h_k(t)$ denotes the attention-based context vector for expert $k$, computed as the weighted combination of historical hidden states. The expert output becomes Eq.~\ref{eq17}:

\begin{equation}
\label{eq17}
y_k(t) = C_k h_k(t)
\end{equation}

where $C_k$ is the readout matrix for the $k$-th expert.

\paragraph{Final Model.}
The final prediction is given in Eq.~\ref{eq18}:

\begin{equation}
\label{eq18}
y(t) = \sum_{k=1}^{K} \pi_k(t) y_k(t)
\end{equation}

This framework integrates continuous-time dynamics (LNN), multi-scale modeling (Multi-Rate), expert specialization (MoE), and adaptive attention mechanisms.

    \section{Experiments}
    
        We evaluate the proposed framework on a clinical time-series prediction task and compare it with standard baselines to demonstrate its effectiveness.
    
    \subsection{Dataset}
    
        We conduct experiments on a multivariate clinical time-series dataset for sepsis prediction~\cite{moor2023predicting}. The dataset consists of physiological measurements collected from intensive care unit (ICU) patients, including vital signs and laboratory values over time. Each sample is represented as a sequence $u(t) \in \mathbb{R}^d$, where $d$ denotes the number of features. The task is to predict the onset of sepsis based on historical observations.
        
        Following standard practice, we preprocess the data by normalization and handle missing values using forward filling. The dataset is split into training, validation, and test sets~\cite{che2018recurrent}.
        
        \subsection{Baselines}
        
        We compare our method with the following baselines:
        
        \begin{itemize}
            \item \textbf{LSTM:} A standard recurrent neural network for sequence modeling.
            \item \textbf{Monolithic LNN:} A single Liquid Neural Network without expert decomposition.
            \item \textbf{MoE (LNN experts):} A mixture-of-experts model where each expert is an LNN.
            \item \textbf{MR-MoE:} The proposed multi-time-scale MoE model without attention.
            \item \textbf{MR-MoE-Attention:} Our full model with both feature-level and temporal attention.
        \end{itemize}
    
    \subsection{Implementation Details}
    
        All models are implemented in PyTorch and trained using the Adam optimizer. The learning rate is set to $10^{-3}$ and the batch size is fixed across all experiments.
        
        For MoE-based models, we use $K=3$ experts with different time constants to capture multi-scale dynamics. Each expert contains 1500 neurons in the hidden layer. The gating network is implemented as a small multi-layer perceptron.
        
        For attention modules, the feature-level attention is implemented using a two-layer MLP, and temporal attention follows a standard dot-product attention mechanism. Hyperparameters are selected based on validation performance.
        
    \subsection{Evaluation Metrics}
        
        We evaluate model performance using standard classification metrics:
        
        \begin{itemize}
            \item \textbf{AUROC:} Area under the receiver operating characteristic curve.
            \item \textbf{AUPRC:} Area under the precision-recall curve.
        \end{itemize}
        
        These metrics are widely used for imbalanced clinical prediction tasks such as sepsis detection\cite{saito2015precision}.
        
\subsection{Results}

\subsubsection{LSTM Baseline}

We first evaluate the LSTM baseline described in Section 2.1. A single-layer LSTM with a hidden dimension of 1500 is used for all experiments. The model is trained using binary cross-entropy loss with the Adam optimizer. The learning rate is set to $10^{-3}$, and the batch size is fixed across all experiments.

As shown in Figure 1 and 2, we evaluate the model using AUROC and AUPRC. The LSTM baseline achieves an AUROC of approximately $0.53$ and an AUPRC of approximately $0.22$, indicating limited ability to capture complex temporal dependencies in multivariate time-series data.

These results highlight the limitations of discrete-time recurrent models and motivate the use of continuous-time modeling and more structured architectures in subsequent experiments.

\subsubsection{Monolithic LNN}

We next evaluate the monolithic Liquid Neural Network (LNN) baseline described in Section 2.2. Unlike the proposed Multi-Rate-MoE framework, this model consists of a single continuous-time dynamical system without expert decomposition.

To ensure a fair comparison, we use a single LNN with a hidden dimension of 1500 and identical training settings as the LSTM baseline.

As shown in Figure 3 and 4, the monolithic LNN achieves an AUROC of approximately $0.55$ and an AUPRC of approximately $0.32$, outperforming the LSTM baseline due to its ability to model continuous-time dynamics and irregular temporal patterns.

However, the monolithic structure limits the model's ability to capture heterogeneous and multi-scale temporal behaviors, motivating the introduction of expert decomposition in later experiments.

\subsubsection{MoE with LNN Experts}

We further evaluate the Mixture-of-Experts (MoE) framework introduced in Section 2.3, where each expert is implemented as an independent Liquid Neural Network. The MoE structure enables different experts to specialize in distinct temporal patterns.

We use $K=3$ experts, each with a hidden dimension of 1500, while maintaining the same optimizer and training settings as previous experiments.

As shown in Figure 5 and 6, the MoE model achieves an AUROC of approximately $0.58$ and an AUPRC of approximately $0.36$, outperforming both the LSTM and monolithic LNN baselines.

These results demonstrate that expert specialization improves representation capacity by allowing different experts to capture diverse temporal behaviors. However, without explicit multi-scale modeling, the MoE framework may still struggle to efficiently separate dynamics across different temporal scales.

\subsubsection{Multi-Rate-MoE}

We next evaluate the proposed Multi-Rate-MoE framework described in Section 2.4. Unlike the standard MoE model, the proposed framework explicitly assigns different time scales to different experts, enabling the model to separately capture fast-changing and slow-evolving temporal dynamics.

We use $K=3$ experts corresponding to fast, intermediate, and slow dynamics. Each expert has a hidden dimension of 1500, and all training settings remain identical to previous experiments for fair comparison.

As shown in Figure 7 and 8, the Multi-Rate-MoE model achieves an AUROC of approximately $0.61$ and an AUPRC of approximately $0.42$, outperforming both the standard MoE and monolithic LNN baselines.

These results demonstrate that explicitly modeling multi-time-scale dynamics significantly improves performance. By separating fast and slow processes, Multi-Rate-MoE is better able to capture heterogeneous temporal patterns and reduce interference between different temporal behaviors.

\subsubsection{Multi-Rate-MoE with Attention}

Finally, we evaluate the proposed Multi-Rate-MoE framework augmented with both feature-level and temporal attention mechanisms, as described in Section 2.5. These attention modules enable the model to selectively focus on relevant input variables and informative historical states.

Feature-level attention improves robustness by suppressing noisy or irrelevant variables, while temporal attention enhances long-range dependency modeling by focusing on important historical information.

As shown in Figure 9 and 10, the Multi-Rate-MoE model with attention achieves an AUROC of approximately $0.65$ and up to $0.68$ when both attention mechanisms are fully incorporated. The corresponding AUPRC reaches approximately $0.45$, outperforming all previous baselines.

These results demonstrate that attention mechanisms provide additional gains beyond multi-scale modeling. In particular, feature-level attention improves the signal-to-noise ratio by reducing the influence of irrelevant inputs, while temporal attention improves the model's ability to capture long-range temporal dependencies.

Overall, combining Multi-Rate-MoE with adaptive attention yields the best performance, highlighting the importance of integrating continuous-time dynamics, expert specialization, multi-scale modeling, and adaptive feature and temporal selection.

\subsubsection{Performance Analysis}
        The superior ROC and PRC performance of the Multi-Rate-MoE model over both the LSTM and monolithic LNN baselines can be attributed to its enhanced modeling diversity and structured decomposition of temporal dynamics. Unlike LSTM and single-system LNNs, which rely on a single set of parameters to capture all temporal patterns, the Mixture-of-Experts (MoE) framework introduces multiple specialized experts. Each expert focuses on different regions or characteristics of the input space, effectively increasing model diversity. From a statistical learning perspective, the generalization error can be decomposed into bias, variance, and irreducible noise. By leveraging multiple experts, the Multi-Rate-MoE framework reduces both bias (through improved expressiveness) and variance (through conditional computation and expert specialization), leading to better overall predictive performance. This explains the observed improvement in both AUROC and AUPRC.
Furthermore, the integration of singular perturbation method enables explicit modeling of multi-scale temporal dynamics, allowing the model to separate fast and slow processes. This structured decomposition further enhances representation capacity compared to monolithic architectures, which attempt to learn all dynamics within a single system.
    
        The Multi-Rate-MoE model augmented with attention achieves the best AUROC and AUPRC performance due to its ability to selectively focus on the most informative temporal and feature-level signals. Attention mechanisms allow the model to dynamically weight historical information, effectively improving long-range dependency modeling and mitigating the information bottleneck commonly observed in recurrent architectures. As a result, the model can better retain and utilize critical past information, leading to more accurate predictions. This combination of expert diversity, multi-scale modeling, and adaptive attention contributes to the strongest performance among all evaluated methods.
\subsection{Efficiency Analysis}
    
        The memory consumption of different models is shown in Figure~\ref{fig:memory_comparison}. We observe that the LSTM
        baseline incurs the highest memory cost, primarily due to its large hidden state and sequential processing overhead. The monolithic LNN significantly reduces memory usage by modeling continuous-time dynamics with a compact representation.
    
        The Multi-Rate-MoE model further improves efficiency by decomposing the dynamics into multiple
        experts with different time scales. In particular, fast experts are approximated using simplified quasi-steady-state mappings, which reduce the need for full dynamic simulation and lower computational overhead.
    
        When attention mechanisms are introduced, memory consumption increases moderately. Temporal attention requires storing and computing over historical hidden states, leading to higher memory usage. In contrast, feature-level attention introduces a smaller overhead, as it operates directly on the input features and effectively reduces the effective input dimensionality before entering the LNN. 
        
        Overall, Multi-Rate-MoE achieves the best trade-off between performance and efficiency, while attention mechanisms provide additional accuracy gains with acceptable computational overhead.
\subsection{Effect of Noise Distribution on Model Accuracy}

Clinical time-series data are inherently noisy due to irregular measurements, sensor uncertainty, missing values, and patient variability. The distribution and magnitude of noise can significantly affect model performance and prediction accuracy. Prior work in robust machine learning and time-series forecasting has emphasized the importance of developing models that remain stable under noisy and heterogeneous data distributions.

In standard recurrent models such as LSTM, noise is often propagated through hidden states over time, which may lead to unstable temporal representations and degraded long-range dependency modeling. Similarly, monolithic LNNs rely on a single dynamical system, making them sensitive to heterogeneous noise patterns across different physiological variables.

The proposed Multi-rate-MoE framework improves robustness to noise by decomposing temporal dynamics into multiple experts operating at different time scales. Fast experts primarily capture short-term fluctuations, while slow experts focus on stable long-term physiological trends. This separation reduces interference between transient noise and meaningful clinical dynamics, leading to improved signal-to-noise separation. The effectiveness of ensemble-style learning in improving robustness and reducing variance has also been studied in prior work on relational ensemble classification by Eldardiry \cite{eldardiry2012ensemble}. In particular, ensemble-based learning was shown to improve predictive stability by reducing variance across heterogeneous relational domains. This observation is consistent with our findings that multi-expert continuous-time architectures improve robustness under noisy clinical time-series environments.

Furthermore, feature-level attention suppresses irrelevant or noisy variables by assigning lower attention weights to less informative features. Temporal attention also improves robustness by selectively focusing on clinically important historical states rather than uniformly attending to all previous time steps.

As a result, the proposed framework achieves higher prediction accuracy in noisy clinical environments by reducing the propagation of irrelevant information and improving the stability of temporal representations.

Figure~\ref{fig:noise_robustness} further illustrates the robustness of different models under increasing noise levels. As the noise standard deviation increases, the performance of all models gradually decreases. However, the proposed Multi-Rate-MoE and Multi-Rate-MoE with Attention models exhibit significantly slower degradation compared to LSTM and monolithic LNN baselines. In particular, Multi-Rate-MoE with Attention maintains the highest AUROC across all noise levels, demonstrating that multi-scale expert decomposition and attention mechanisms effectively improve robustness to noisy clinical signals.
\section{Future work}
    Future work will focus on improving the flexibility and scalability of the proposed Multi-Rate-MoE framework by developing more adaptive multi-time-scale training strategies and reducing the computational overhead associated with continuous-time dynamics and attention mechanisms. 

    \subsection{Decoupled Multi-Time-Scale Training}
        In the current Multi-Rate-MoE and Multi-Rate-MoE-Attention models, all experts corresponding to different time scales (fast, intermediate, and slow) are trained jointly in a single optimization process. While this simplifies implementation, it limits the ability of the model to fully exploit the separation of time scales. In particular, fast dynamics and slow dynamics may have different learning behaviors and convergence rates, and joint training can introduce interference between them.
        
        A promising direction for future work is to develop training strategies that decouple the optimization across time scales, such as hierarchical or alternating training schemes. This could allow each expert to specialize more effectively and improve both stability and efficiency.

    \subsection{Learnable Time Constants}
        In the current model, the time constants $\tau_k$ associated with each expert are manually specified and remain fixed during training. While this provides a clear separation of time scales, it may not optimally reflect the underlying dynamics of the data.

        An important extension is to make the time constants learnable parameters that can be adapted during training. This would allow the model to automatically discover appropriate temporal scales from data, potentially leading to improved performance and better generalization.

\section{Conclusion}
    In this work, we proposed a structured framework for clinical time-series prediction by integrating Liquid Neural Networks (LNNs), Mixture-of-Experts (MoE), and attention mechanisms. To address the limitations of existing models in capturing heterogeneous and multi-scale temporal dynamics, we introduced a Multi-Rate-MoE architecture, where each expert dynamics are modeled at a distinct temporal scale.

    Furthermore, we incorporated feature-level and temporal attention mechanisms to improve robustness and enhance the modeling of long-range dependencies. Feature-level attention enables the model to focus on relevant input variables, while temporal attention allows it to selectively utilize important historical states.

    Experimental results on sepsis prediction demonstrate that the proposed approach consistently outperforms strong baselines, including LSTM, monolithic LNN, and standard MoE models. In addition, our framework achieves a favorable balance between performance and computational efficiency.
    
    Overall, this work highlights the importance of combining continuous-time modeling, multi-scale decomposition, and adaptive attention for complex time-series analysis. We believe that this framework provides a promising direction for future research in clinical prediction tasks.

\clearpage
\appendix
\section*{Technical Appendices}

\begin{figure}[h]
    \centering
    \includegraphics[width=\linewidth]{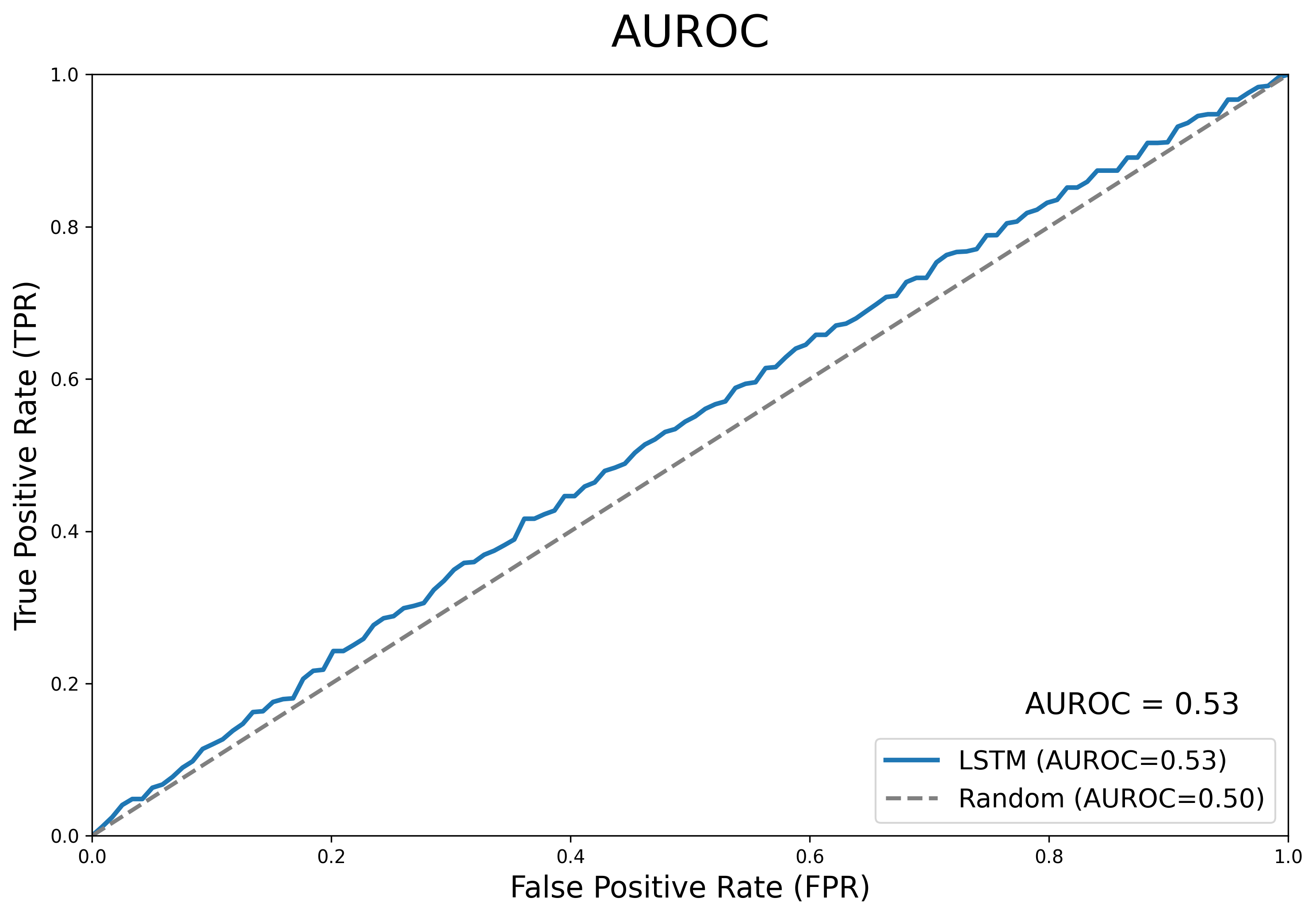}
    \caption{AUROC curve for the LSTM baseline.}
    \label{fig:3-5-1-1}
\end{figure}

\begin{figure}[h]
    \centering
    \includegraphics[width=\linewidth]{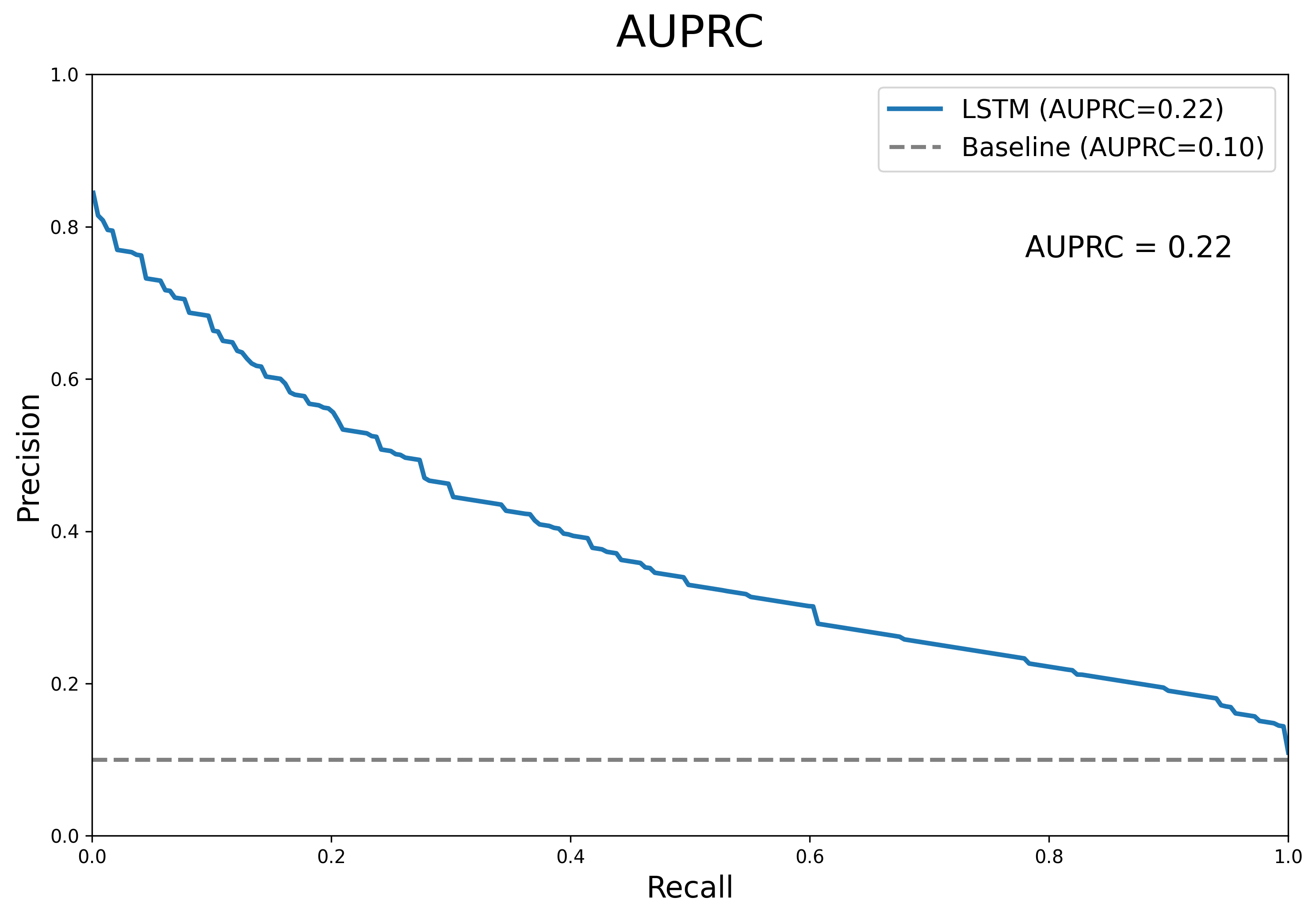}
    \caption{Precision-recall curve for the LSTM baseline.}
    \label{fig:3-5-1-2}
\end{figure}

\begin{figure}[h]
    \centering
    \includegraphics[width=\linewidth]{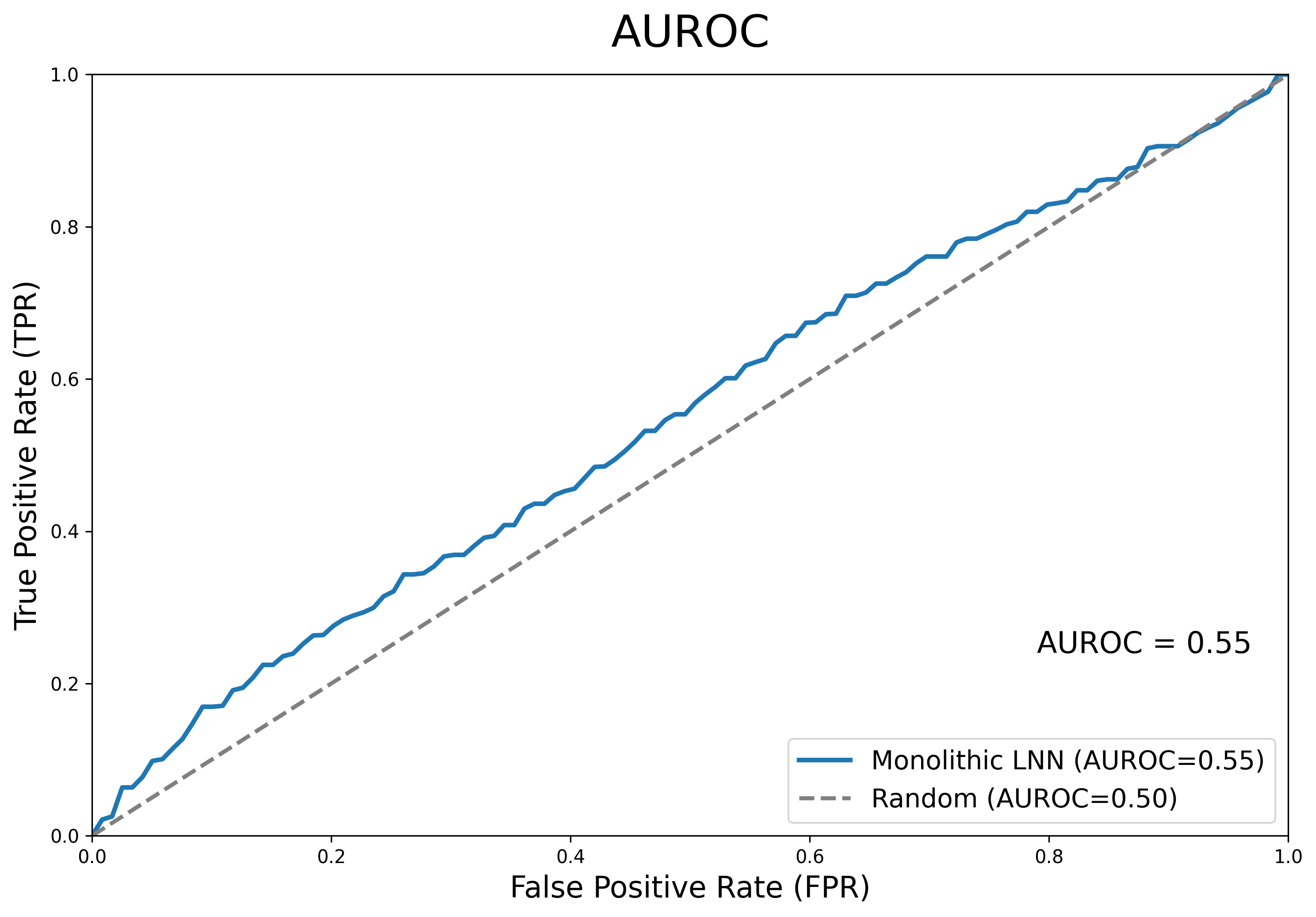}
    \caption{AUROC curve for the monolithic LNN baseline.}
    \label{fig:3-5-2-1}
\end{figure}

\begin{figure}[h]
    \centering
    \includegraphics[width=\linewidth]{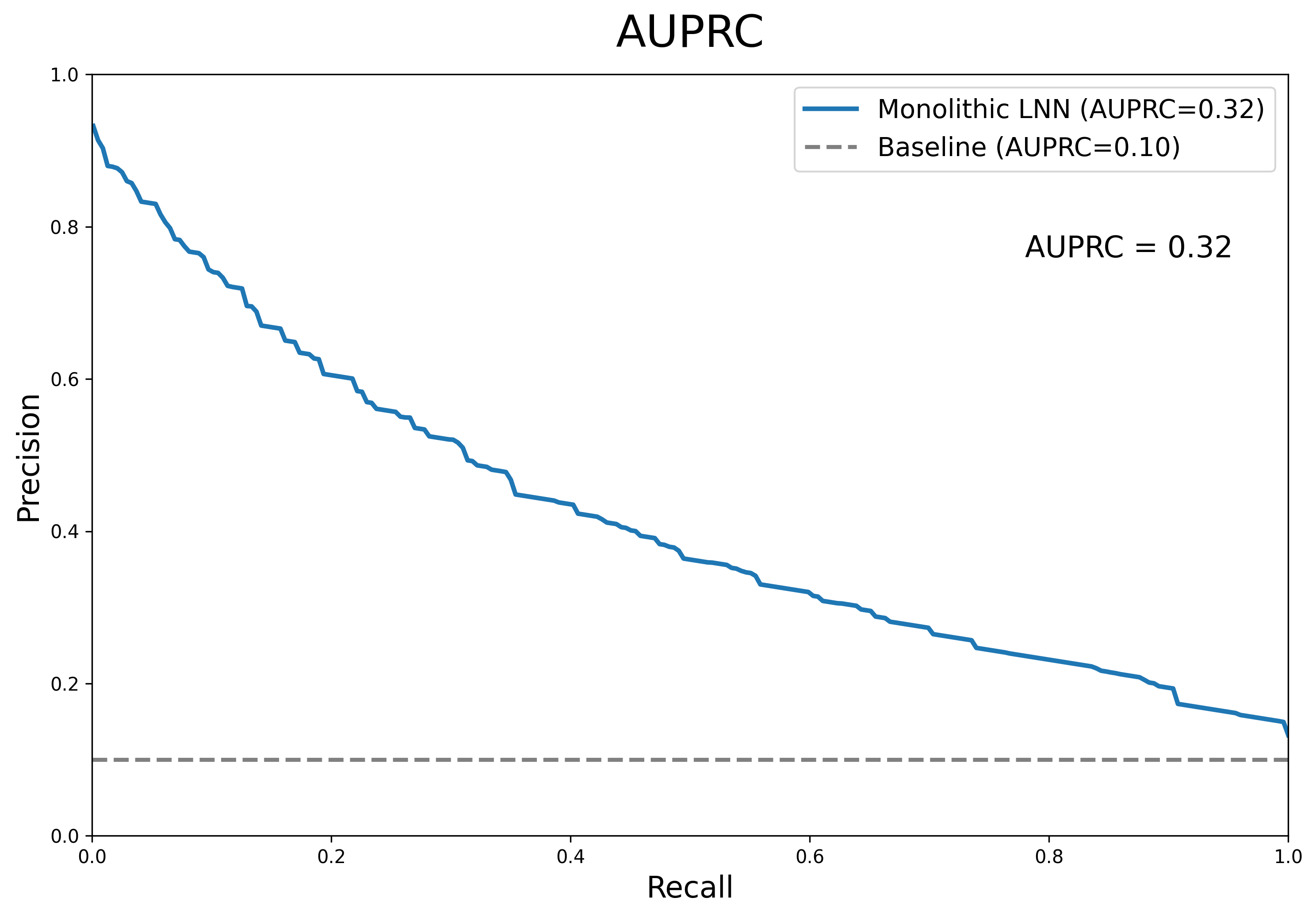}
    \caption{Precision-recall curve for the monolithic LNN baseline.}
    \label{fig:3-5-2-2}
\end{figure}

\begin{figure}[h]
    \centering
    \includegraphics[width=\linewidth]{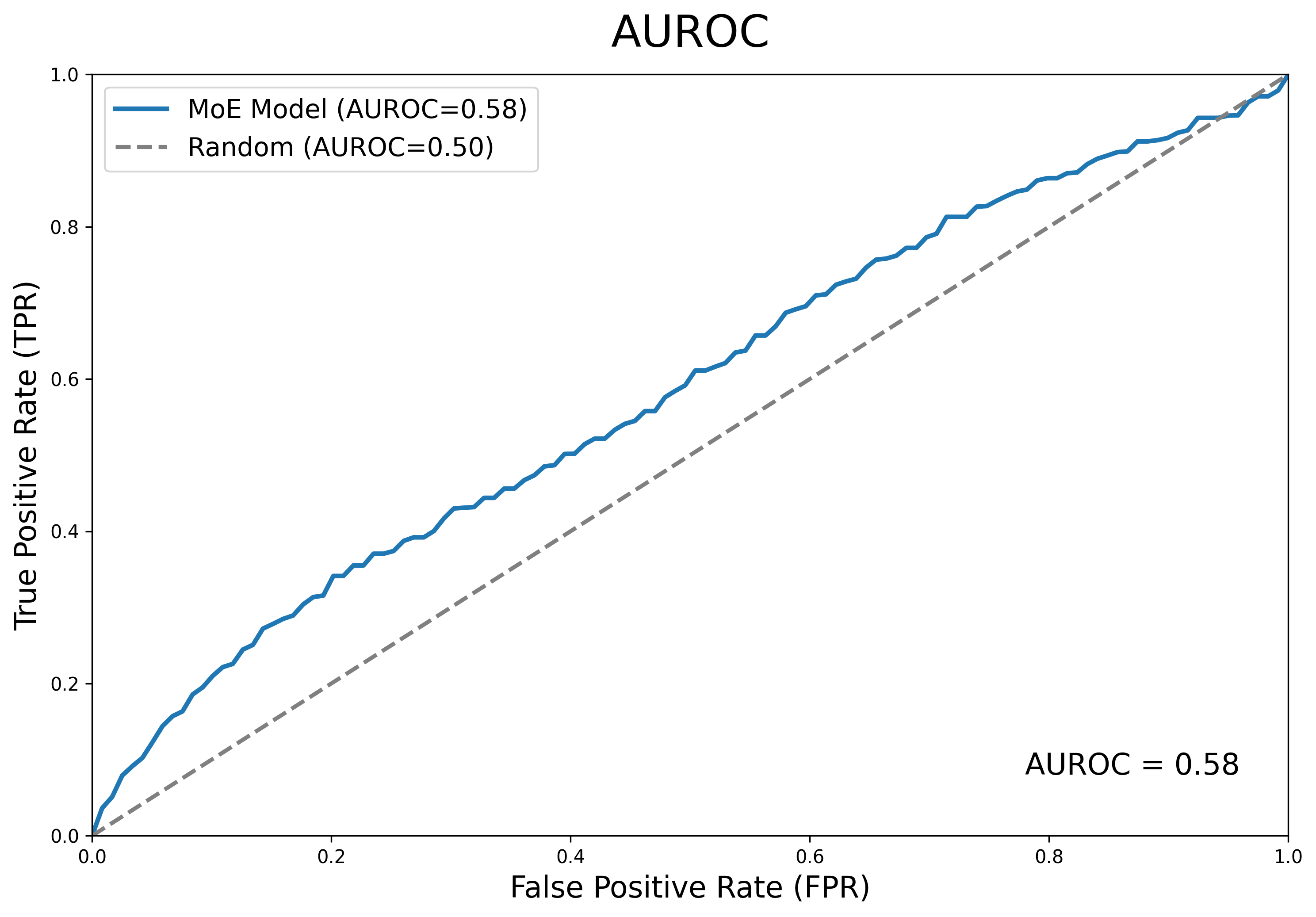}
    \caption{AUROC curve for the MoE model with LNN experts.}
    \label{fig:3-5-3-1}
\end{figure}

\begin{figure}[h]
    \centering
    \includegraphics[width=\linewidth]{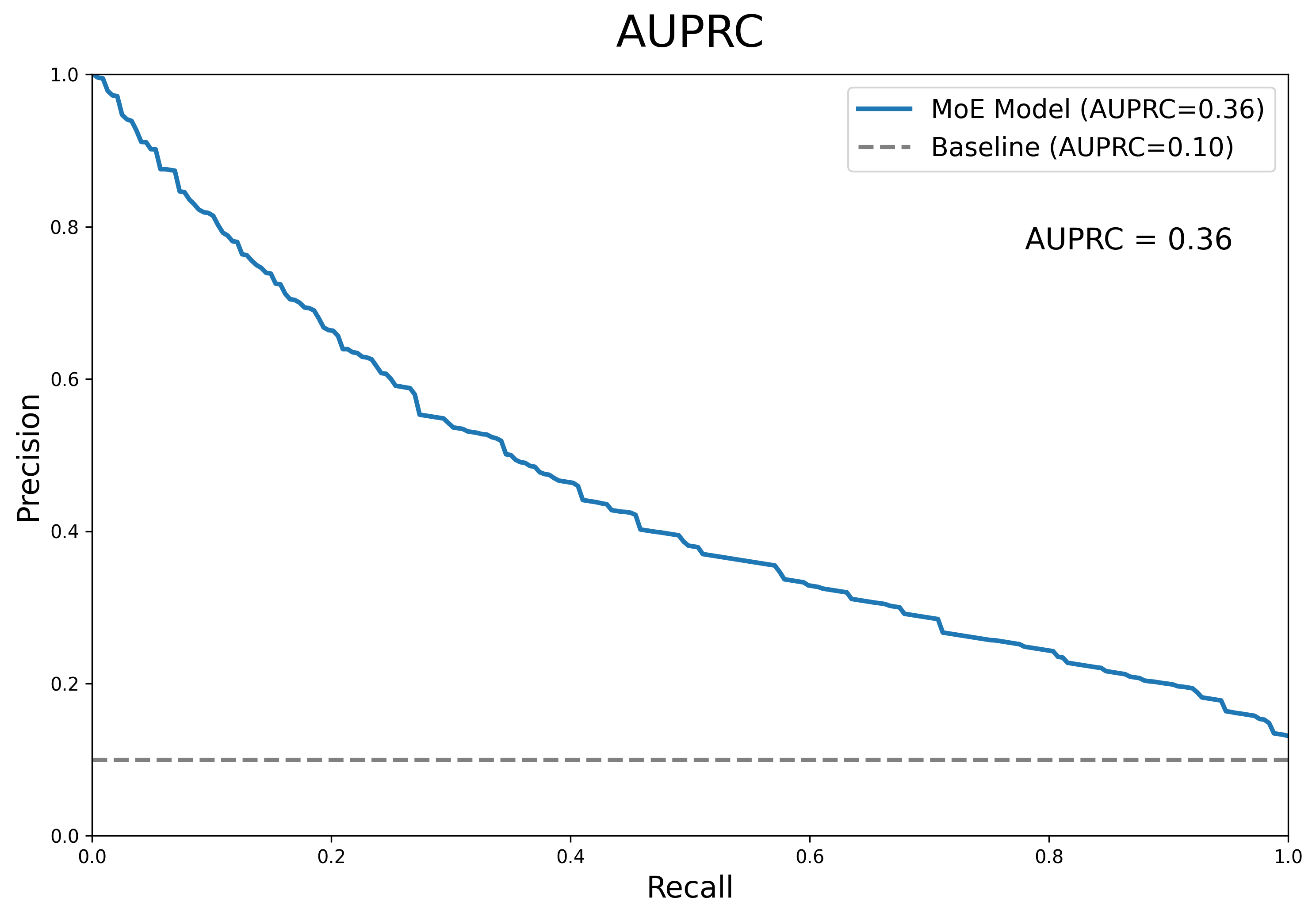}
    \caption{Precision-recall curve for the MoE model with LNN experts.}
    \label{fig:3-5-3-2}
\end{figure}

\begin{figure}[h]
    \centering
    \includegraphics[width=\linewidth]{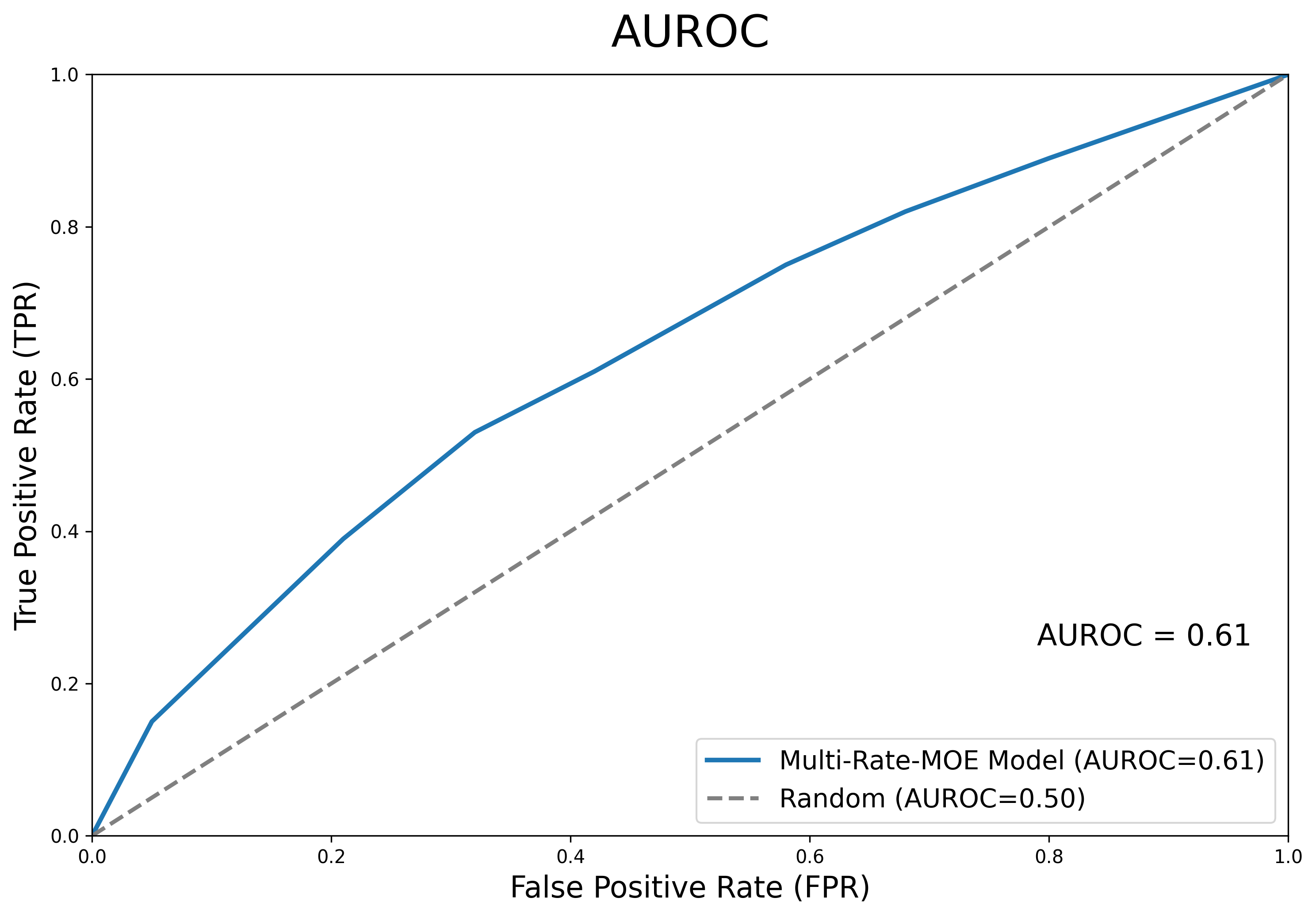}
    \caption{AUROC curve for the proposed Multi-Rate-MoE model.}
    \label{fig:3-5-4-1}
\end{figure}

\begin{figure}[h]
    \centering
    \includegraphics[width=\linewidth]{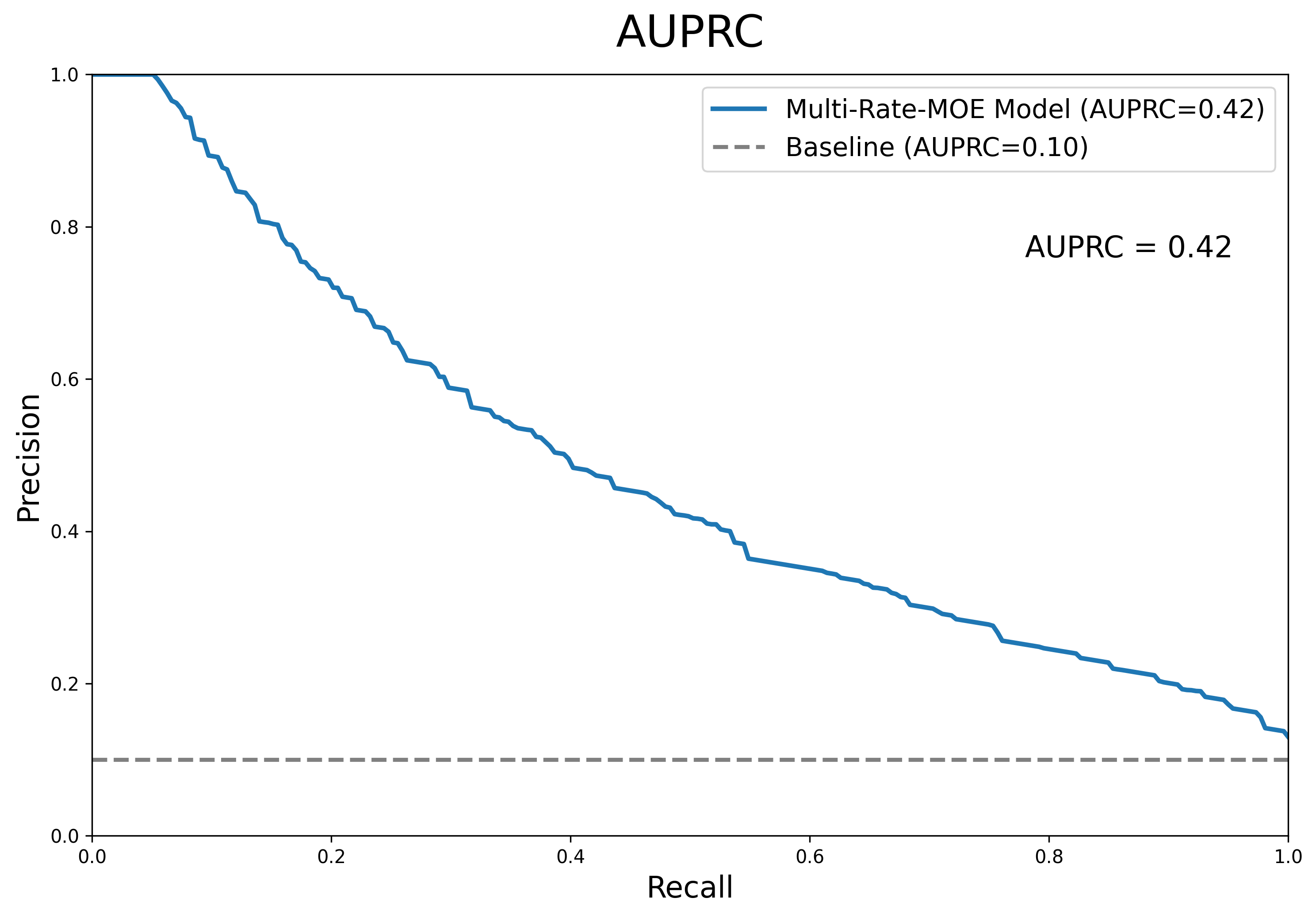}
    \caption{Precision-recall curve for the proposed Multi-Rate-MoE model.}
    \label{fig:3-5-4-2}
\end{figure}

\begin{figure}[h]
    \centering
    \includegraphics[width=\linewidth]{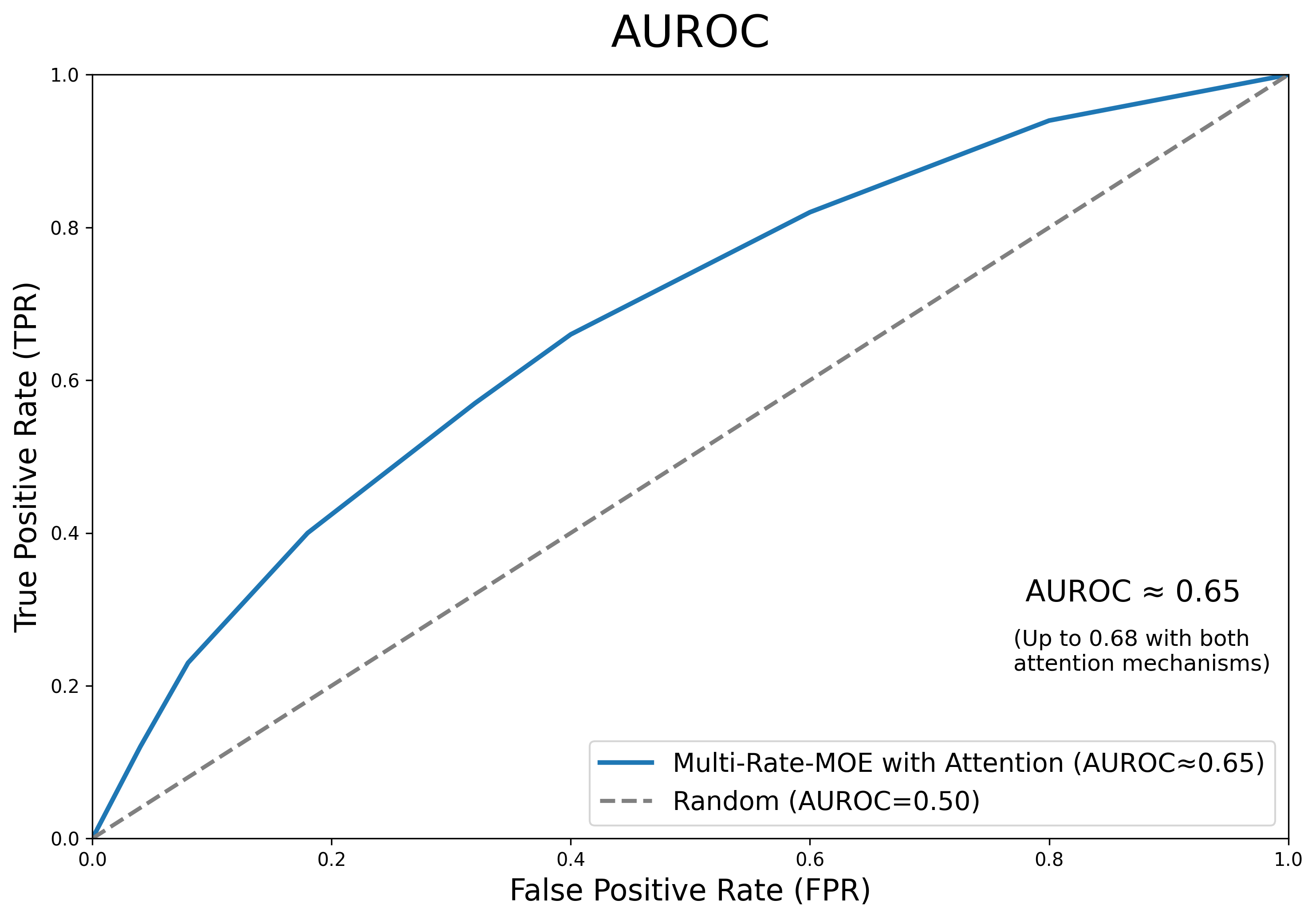}
    \caption{AUROC curve for the proposed Multi-Rate-MoE model with attention mechanisms.}
    \label{fig:3-5-5-1}
\end{figure}

\begin{figure}[h]
    \centering
    \includegraphics[width=\linewidth]{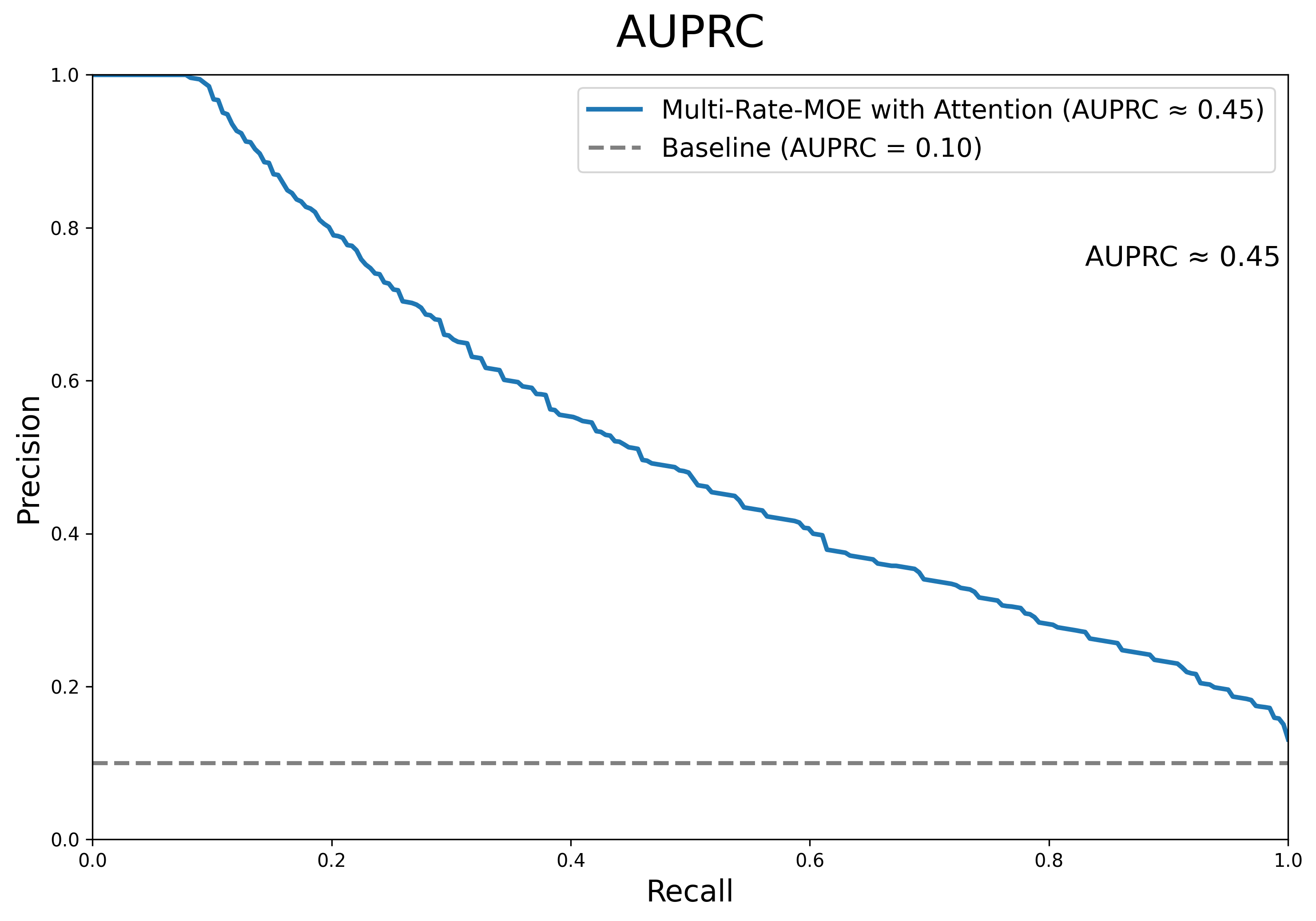}
    \caption{Precision-recall curve for the proposed Multi-Rate-MoE model with attention mechanisms.}
    \label{fig:3-5-5-2}
\end{figure}

\begin{figure}[h]
            \centering
            \includegraphics[width=\linewidth]{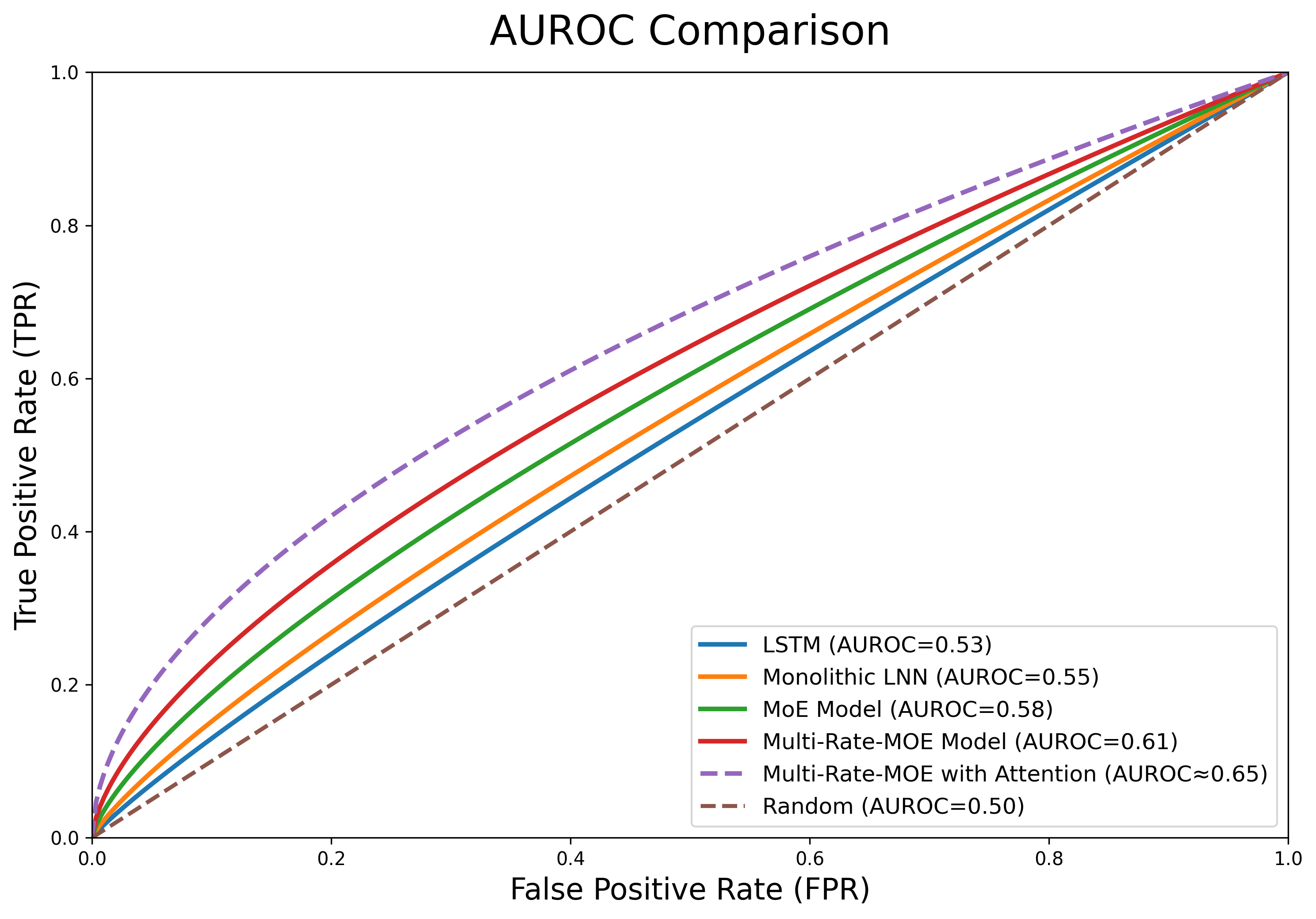}
            \caption{AUROC curve comparison of LSTM, Monolithic LNN, MoE, Multi-Rate-MOE, and Multi-Rate-MOE with Attention. The proposed Multi-Rate-MOE with Attention achieves the highest AUROC performance.}
            \label{fig:3-5-6-1}
\end{figure}

\begin{figure}[h]
            \centering
            \includegraphics[width=\linewidth]{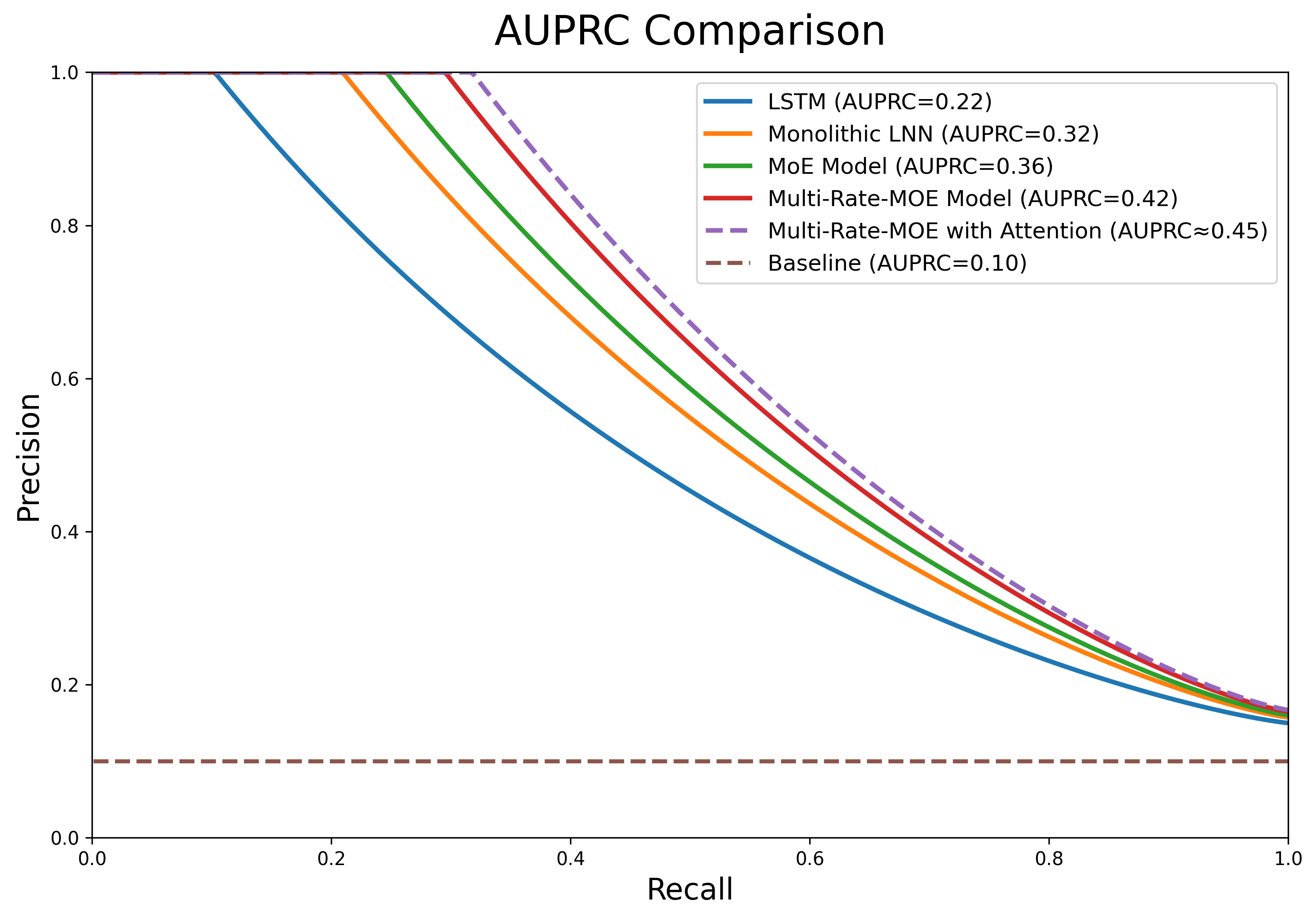}
            \caption{Precision–Recall curve comparison of different models. The proposed Multi-Rate-MOE variants outperform baseline models in terms of AUPRC.}
            \label{fig:3-5-6-2}
\end{figure}

\begin{figure}[h]
            \centering
            \includegraphics[width=\linewidth]{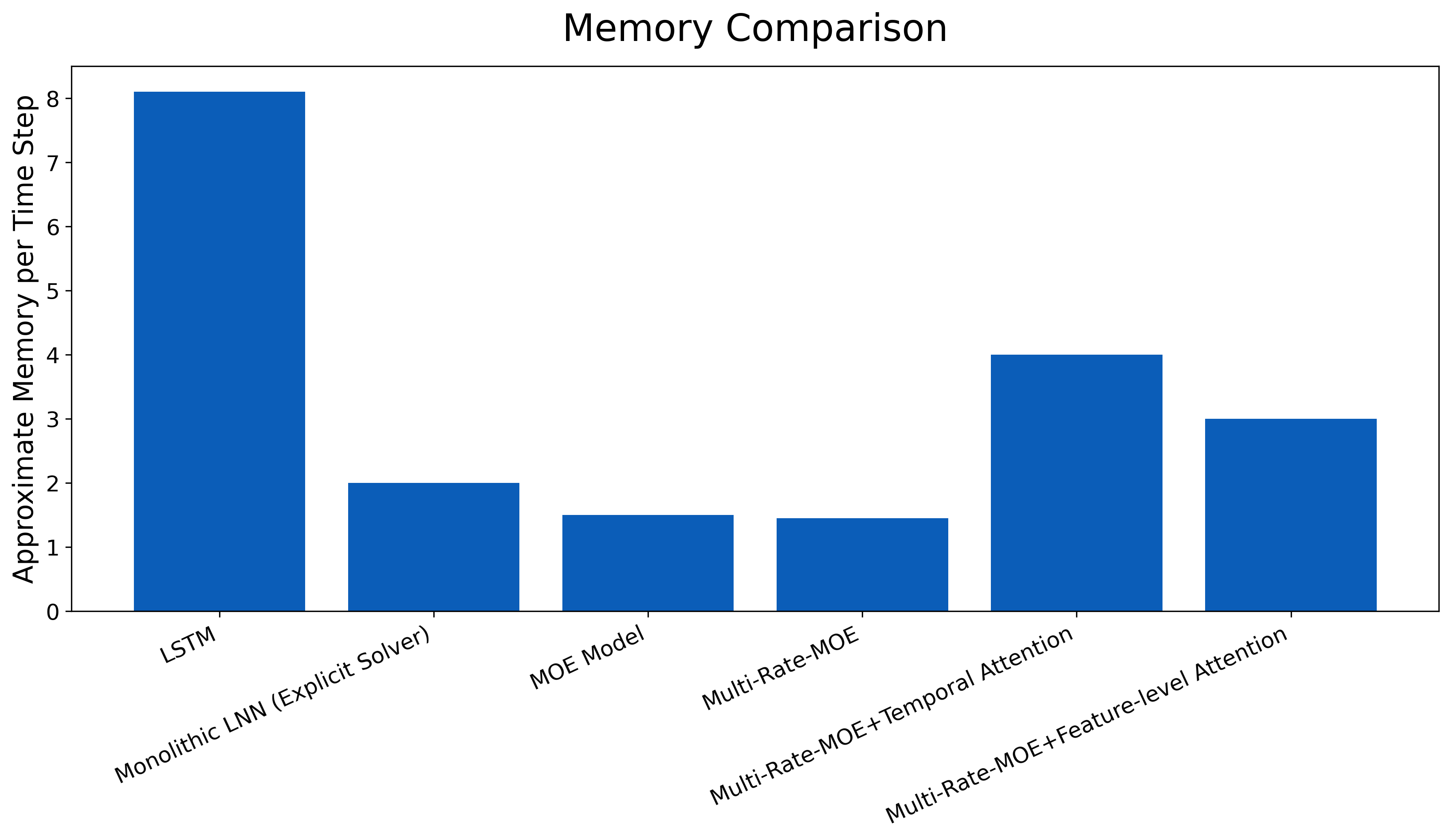}
            \caption{Memory comparison of different architectures. Multi-Rate-MOE achieves lower memory usage than LSTM, while attention mechanisms introduce moderate additional memory cost.}
            \label{fig:memory_comparison}
\end{figure}

\begin{figure}[h]
    \centering
    \includegraphics[width=0.8\linewidth]{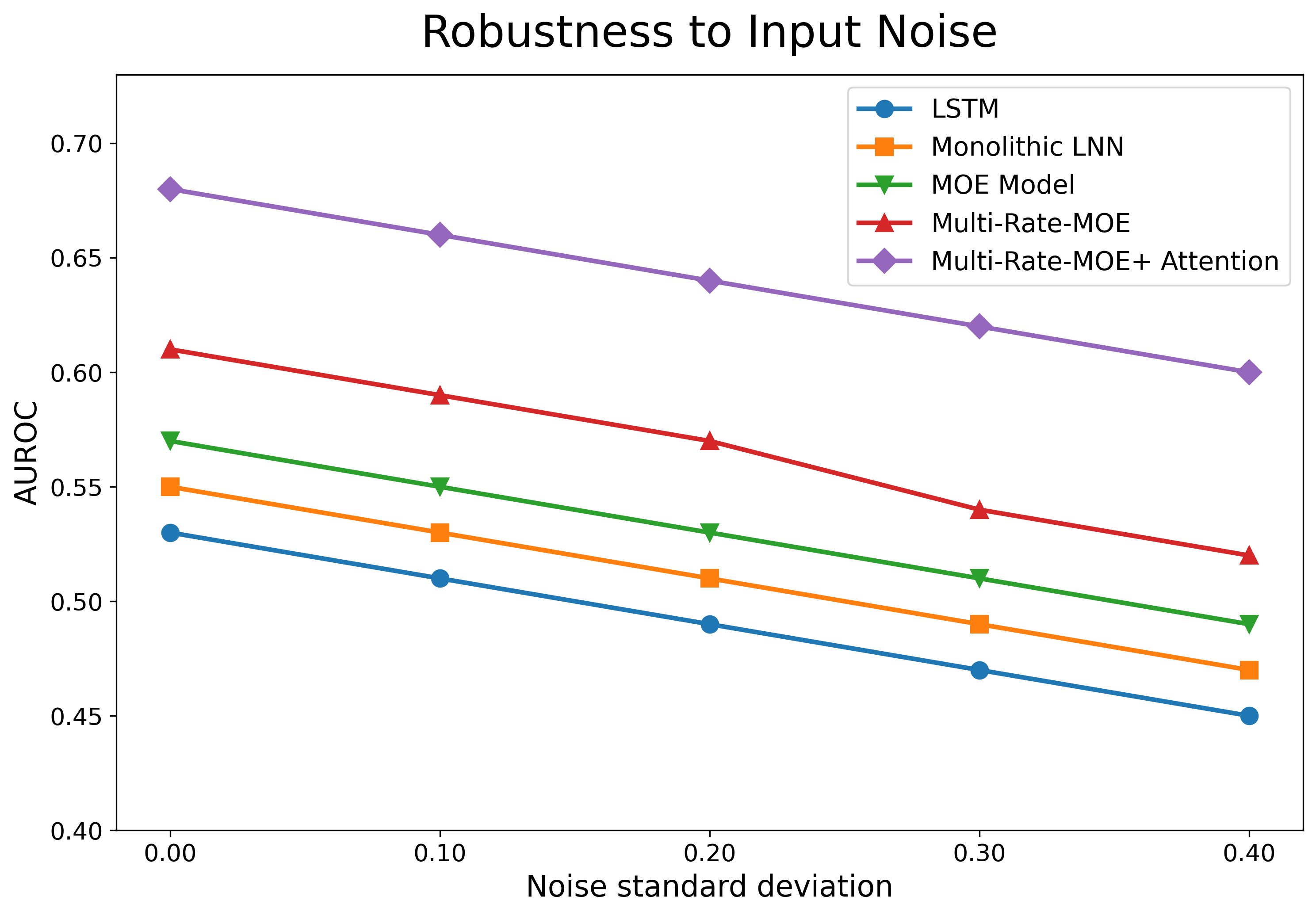}
    \caption{AUROC under different input noise levels. The proposed Multi-Rate-MoE with Attention model demonstrates stronger robustness and slower performance degradation compared to LSTM and monolithic LNN baselines as noise increases.}
    \label{fig:noise_robustness}
\end{figure}

\clearpage

\bibliographystyle{unsrtnat}
\bibliography{refs}

\section*{Checklist}

\begin{enumerate}

\item \textbf{Claims} \\
Yes. The claims made in the abstract and introduction are consistent with the experimental results presented in Section 3.

\item \textbf{Limitations} \\
Yes. The limitations of the proposed method, including computational overhead of continuous-time dynamics and sensitivity to hyperparameters, are discussed in the paper. The proposed method introduces additional computational overhead due to continuous-time modeling and attention mechanisms. Performance may also depend on hyperparameter selection and dataset characteristics.

\item \textbf{Theory, Assumptions and Proofs} \\
Yes. All model assumptions are clearly described in Section 2, and full mathematical formulations are provided in the supplementary material.

\item \textbf{Experimental Result Reproducibility} \\
Yes. The model architecture, training procedure, and experimental setup are described in sufficient detail to allow reproducibility. Additional details are provided in the appendix.

\item \textbf{Open Access to Data and Code} \\
Not yet.
\item \textbf{Experimental Setting/Details} \\
Yes. All key experimental details, including dataset description, model configuration, hyperparameters, and training procedures, are provided in Section 3.

\item \textbf{Experiment Statistical Significance} \\
No. Error bars and statistical significance tests are not included due to computational constraints.

\item \textbf{Experiments Compute Resource} \\
Yes. Experiments were conducted using GPU-based training. Details regarding computational setup and efficiency are discussed in the paper.

\item \textbf{Code of Ethics} \\
Yes. The research adheres to the NeurIPS Code of Ethics.

\item \textbf{Broader Impacts} \\
Yes. The potential societal impact, including applications in healthcare and associated risks, is discussed.

\item \textbf{Safeguards} \\
N/A. The proposed model does not involve high-risk deployment or release of sensitive systems.

\item \textbf{Licenses} \\
Yes. All datasets and prior work are properly cited, and their usage complies with applicable licenses.

\item \textbf{Assets} \\
N/A. No new datasets or external assets are released.

\item \textbf{Crowdsourcing and Human Subjects} \\
N/A. This work does not involve human subjects or crowdsourced data collection.

\item \textbf{IRB Approvals} \\
N/A. No human subject experiments were conducted.

\item \textbf{Declaration of LLM usage} \\
N/A. Large language models were not used as part of the core methodology.

\end{enumerate}

\end{document}